\documentclass[runningheads]{llncs}
\pdfoutput=1

\usepackage{graphicx}

\usepackage{tikz}
\usepackage{comment}
\usepackage{amsmath,amssymb} 
\usepackage{color}

\usepackage[accsupp]{axessibility}  

\usepackage{graphicx}
\usepackage{mathtools}
\usepackage{bbm}
\usepackage{amssymb}
\usepackage{booktabs}
\usepackage{multirow}
\usepackage[utf8]{inputenc}
\newcommand{\norm}[1]{\left\lVert#1\right\rVert_2}

\newcommand{\normfro}[1]{\left\lVert#1\right\rVert_\text{F}}
\newcommand{\cmark}{\ding{51}}%
\newcommand{\xmark}{\ding{55}}%
\usepackage{pifont}
\usepackage{listings}
\usepackage{thmtools, thm-restate}

\newcommand{\myparagraph}[1]{\vspace{3pt}\noindent{\bf #1}}

\newcommand{\ours}{MSBReg }

\begin{document}
\pagestyle{headings}
\mainmatter
\def\ECCVSubNumber{7156}  

\title{An Embedding-Dynamic Approach to Self-supervised Learning} 

\titlerunning{An Embedding-Dynamic Approach to Self-supervised Learning}
%
\author{Suhong Moon\inst{1} \and
Domas Buracas\inst{1} \and
Seunghyun Park\inst{2} \and
Jinkyu Kim\inst{3} \and
John Canny\inst{1}}
\authorrunning{S. Moon et al.}
%
\institute{Univerity of California, Berkeley \and NAVER Clova \and Korea University}
\maketitle

\begin{abstract}
A number of recent self-supervised learning methods have shown impressive performance on image classification and other tasks. A somewhat bewildering variety of techniques have been used, not always with a clear understanding of the reasons for their benefits, especially when used in combination. Here we treat the embeddings of images as point particles and consider model optimization as a dynamic process on this system of particles. Our dynamic model combines an attractive force for similar images, a locally dispersive force to avoid local collapse, and a global dispersive force to achieve a globally-homogeneous distribution of particles. The dynamic perspective highlights the advantage of using a delayed-parameter image embedding (a la BYOL) together with multiple views of the same image. It also uses a purely-dynamic local dispersive force (Brownian motion) that shows improved performance over other methods and does not require knowledge of other particle coordinates. The method is called \ours which stands for (i) a $\textbf{M}\text{ultiview}$ centroid loss, which applies an attractive force to pull different image view embeddings toward their centroid, (ii) a $\textbf{S}\text{ingular value}$ loss, which pushes the particle system toward spatially homogeneous density, (iii) a $\textbf{B}\text{rownian}$ diffusive loss. We evaluate downstream classification performance of \ours on ImageNet as well as transfer learning tasks including fine-grained classification, multi-class object classification, object detection, and instance segmentation. In addition, we also show that applying our regularization term to other methods further improves their performance and stabilize the training by preventing a mode collapse.
\end{abstract}

\section{Introduction}\label{sec:introduction}
\begin{figure}[h]
    \begin{center}
        \includegraphics[width=.7\linewidth]{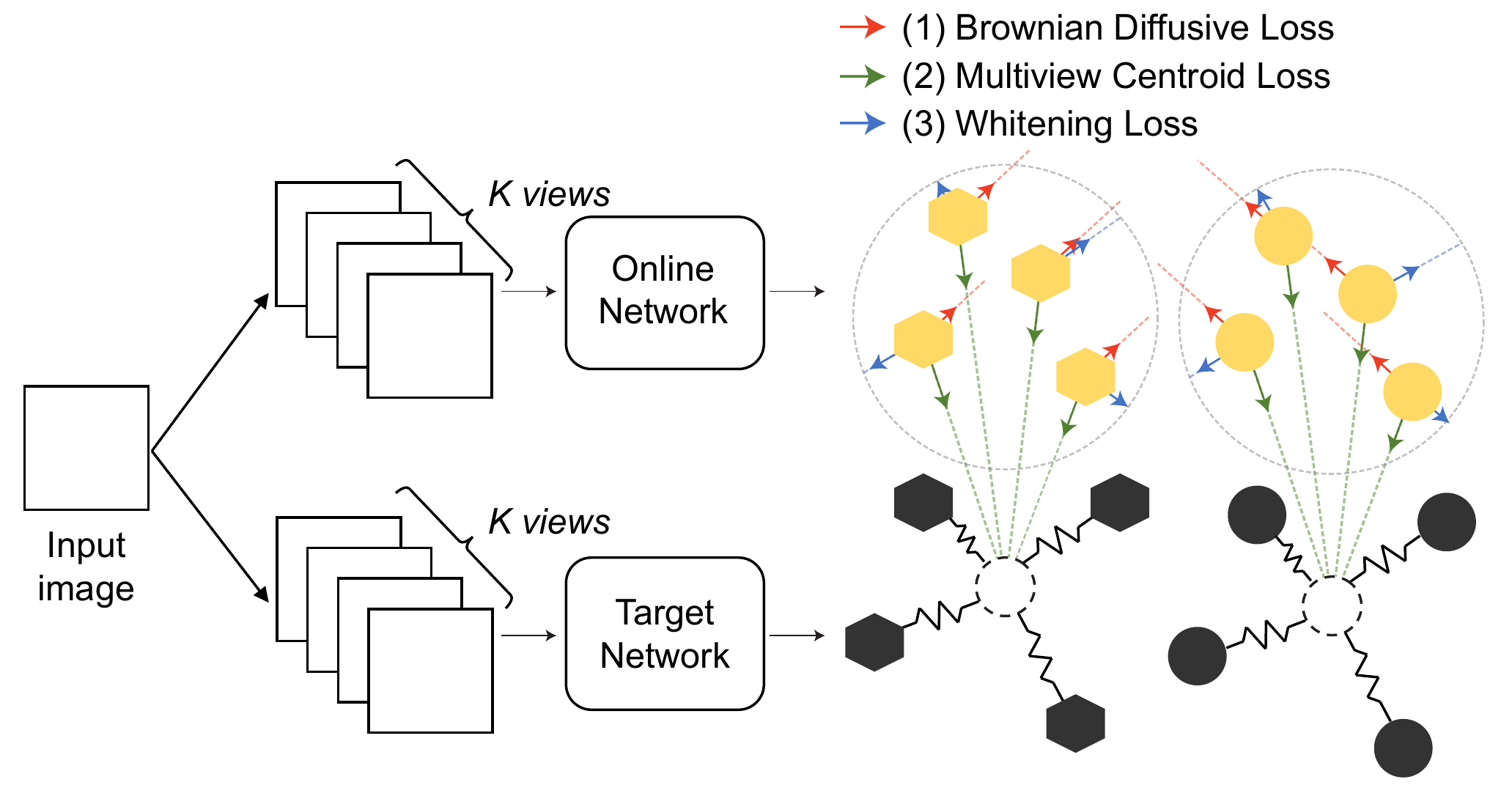}
    \end{center}
    \caption{Our proposed \ours for SSL contains the following three regularization terms. (1) A Brownian diffusive loss (red), which induces a random motion of embeddings. This provides an implicit contrastive effect, preventing collapse of emebddings and stretching weaker links more on average than strong ones. (2) A Multiview centroid loss (green), where we train our online network to minimize the distance between centroids of online and target network embeddings of different views of the same image. (3) A Singular value loss (blue), which decorrelates the different feature dimensions to disperse embeddings uniformly in the embedding space. Positive pairs are indicated with the same shapes.}
    \label{fig:teaser}
    \vspace{-2em}
\end{figure}
A good representation should include useful features (those that facilitate downstream prediction tasks) while ignoring ``nuisance" features~\cite{bengio2013representation}. Among the best known self-supervised methods, contrastive methods combine an attractive term between similar images (typically different perturbations of the same image) with an explicit repulsive term between distinct pairs. Recently, BYOL~\cite{grill2020bootstrap} utilized siamese neural networks (referred to as the online and target) with lagged (moving averaged) parameters in the target network, and simply minimized distance between online and target network embeddings. While there was no {\em explicit} repulsive term in BYOL, it was later shown to be highly dependent on the use of BatchNorm layers. 
The activation normalization in BatchNorm and other methods can be viewed as a global, dimension-wise dispersion of the set of embeddings, a desirable feature of a representation. However, other normalization methods such as LayerNorm were shown to be much less effective in BYOL suggesting the story is more complicated than normalization and global dispersion~\cite{layer_norm,richemond2020byol_wo_bn,untitled-ai,tian2021understanding}. 

Inspection of the gradients in BatchNorm reveals that they have a strong stochastic component (beyond the global ``normalizing" component) that depends on differences between image activations and their batch centroid (i.e. on whatever other images happen to be in the same minibatch). From our perspective, these forces provide a local (stochastic) dispersive force between embeddings. Thus Batchnorm implements two of the desirable features of a good representation (local and global dispersion of embeddings), but in a suboptimal way. Here we define separate loss terms for local and global dispersion and apply them to the embedding layer only (as opposed to other intermediate network layers). By moving dispersion to loss layers, we allow network normalization layers (which ideally impact network training and stability but not losses) to be independently designed. We can also separately define and optimize the local and global dispersion losses. 

If we assume that the optimization method used to train the network is either stateless or sufficiently ``fast'' (e.g. the optimizer uses momentum=$0.9$ for an effective time constant of $10$ steps), then the embeddings are part of a second-order dynamical system. The embeddings are defined by the parameters of the two networks (online and target), together with corresponding input images. The moving parameter average implemented on the target network, together with a fast optimizer which functions as an integrator of loss gradients, defines a second-order dynamical system. We exploit the dynamics of this system in two ways: by using ``fast-slow'' optimization for attractive and dispersive forces, and by showing that stochastic energy injected into the system should ``stretch'' attractive links with equal potential energy on average - so weaker attractive links will be stretched further. 

Multiview contrastive training, where more than two augmentations are compared, works very effectively with a lagged-arm network. Dispersive forces act in the network to situate embeddings with globally uniform density. While loss-gradients act as forces applied to the online network, online embeddings experience a strong viscous ``drag'' from their corresponding target network embeddings which they are attracted to. So embeddings move globally at the time constant of the lagged network, which is typically thousands to tens of thousands of time steps. 
On the other hand, embeddings within the same group, i.e. embeddings of views of the same image, experience no ``drag'' relative to their centroid. They collapse and are maintained close together at the time constant of the online network. 

Given a lagged-arm, siamese architecture inspired by BYOL, we explore Multiview, Singular value, Brownian Diffusive regularizations. These three loss terms address respectively, (i) fast-slow attractive/dispersive optimization, (ii) global, uniform, dispersion of embeddings (iii) local dispersive force. 

We evaluate our approach on visual benchmarks including ImageNet-100~\cite{deng2009imagenet}, STL-10~\cite{stl10}, and ImageNet~\cite{deng2009imagenet}. We show that our model significantly outperforms prior work in the image classification task. Also, we show that joint local/global dispersive forces lead to a larger dissimilarity of negative pairs compared to other approaches. We summarize our contributions as follows:
\begin{itemize}
    \item We analyze and optimize self-supervised learning using a dynamic model of embeddings.\vspace{-0.1em}
    \item To optimize the placement of embeddings, we propose a \ours loss that consists of 1) Multiview centroid loss and 3) Singular value loss 3) Brownian diffusive loss and \ours  outperforms other baselines by a significant margin.
\end{itemize}

\begin{figure*}[t]
    \begin{center}
        \includegraphics[width=\linewidth]{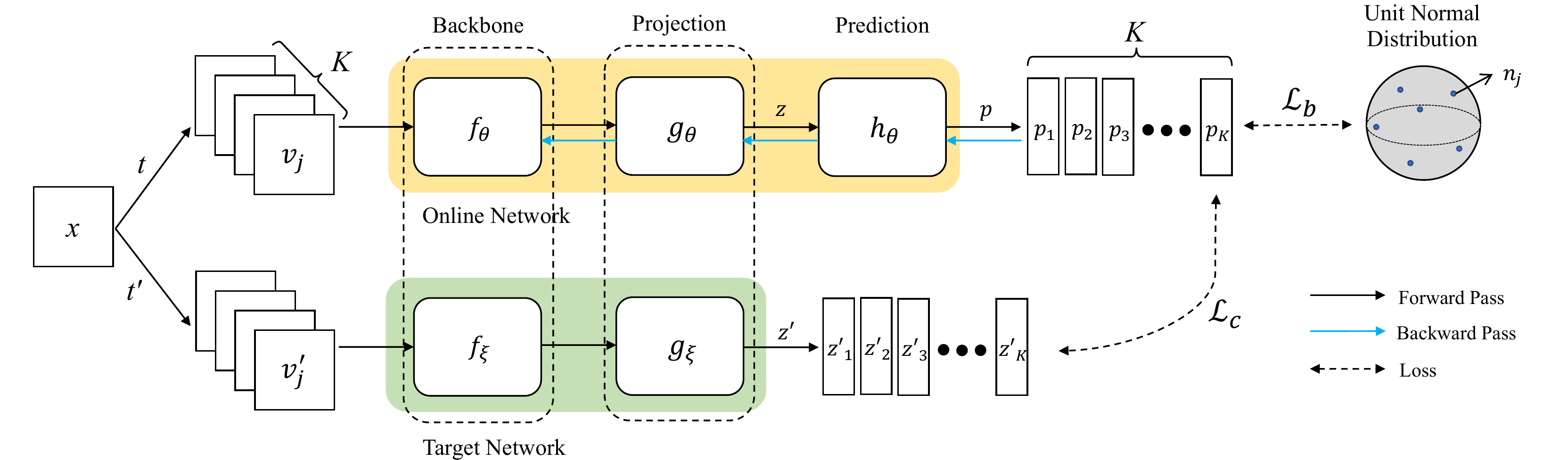}
    \end{center}
    \caption{The architecture overview of \ours. This architecture is inspired by BYOL's architecture. Each model takes K augmented views as its inputs. \ours minimizes (1) multiview centroid loss, (2) singular value loss, (3) Brownian diffusive loss and . The first makes the online network predict the target network's representation of the centroid of K views.  The second loss favors a spatially uniform (spherical) distribution. The last one induces noise into embedding space and makes the embeddings repulse each other on average, preventing the model from converging to collapsed solutions. 
    }
    \label{fig:model_overviews}
    \vspace{-1em}
\end{figure*}

\begin{figure}[t]
    \begin{center}
        \includegraphics[width=\linewidth]{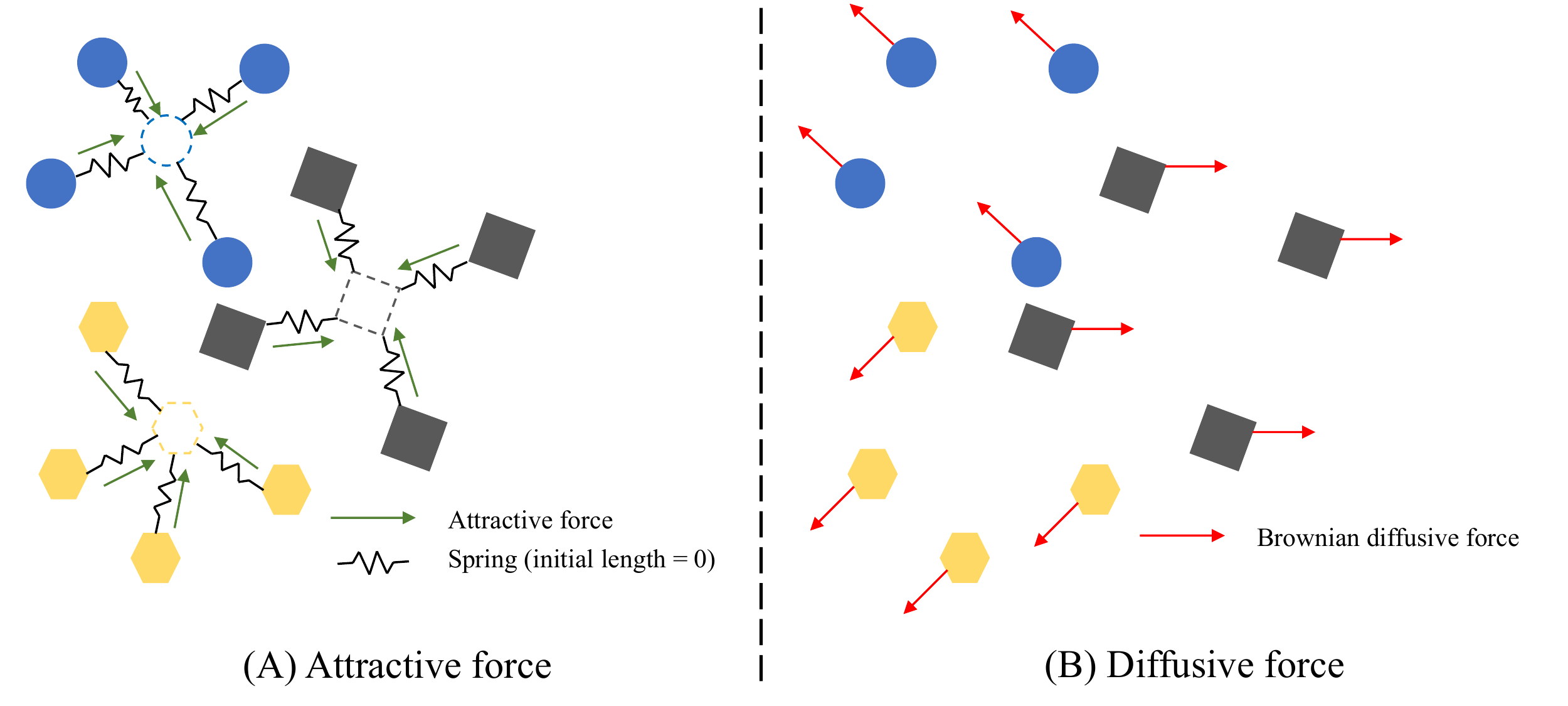}
    \end{center}
    \caption{(A) Multiview centroid loss applies attractive force to embeddings generated by online network. The solid shapes are embeddings generated by online network and the shapes with dashed line are the geometric centroids of the embeddings generated by target network. We can model such system as spring-mass system. (B) Brownian diffusive loss induces the random walk of embeddings, preventing the model from admitting collapsed solutions.}
    \label{fig:explain_on_2d}
    \vspace{-2em}
\end{figure} 

\section{Related Work} \label{sec:related_work}
\myparagraph{Self Supervised Learning.}
Recent works suggest that a state-of-the-art image representation can be successfully and efficiently learned by a discriminative approach with self supervised learning. These methods originally relied on a contrastive loss that encourages representation of different views of the same image (i.e. positive pairs) to be close in the embedding space, and representations of views from different images (i.e. negative pairs) to be pushed away from each other. Contrastive methods often require a careful treatment of negative pairs, which need a large memory overhead as they need to be sampled from a memory bank~\cite{memory_bank,moco_he_2020_cvpr} or from the current mini-batch of data~\cite{simclr}. The contrastive approach is also unsatisfying from a modeling perspective - the fact that images are distinct does not imply that they are different - but the contrastive approad applies large repulsive gradients to distinct, close image pairs. 

Motivated by a desire to overcome the difficulties of contrastive approaches, recent works~\cite{grill2020bootstrap,chen2020simsiam} use two neural networks (referred to as online and target networks) are trained to minimize the distance between their embeddings of the same image. Some works use a moving average on parameters of one arm~\cite{grill2020bootstrap} while others use the same parameters~\cite{moco_he_2020_cvpr,chen2020simsiam}. These methods have been effective but their success is somewhat mysterious since there is no obvious force to prevent collapse of embeddings since forces are only attractive. It turn out that as batch normalization~\cite{batch_norm} was an important element of the success of BYOL. In contrast we employ explicit local and global dispersive losses in addition to attractive forces on groups of multiple image views. 

\myparagraph{Regularizing Consistency of Singular Value.}
Whitening is the most similar approach to regularizing consistency of singular values. Recently, whitening output embeddings has received attention as a method to avoid a mode collapse. Whitening removes redundant information in input and prevents all dimensions of embeddings from being encoded with the same representation. Whitening features induces the contrastive effect of the embeddings by scattering them. This~\cite{whitening_for_self_supervised} performs a explicit whitening via Cholesky decomposition. Performing Cholesky decomposition, which requires the computation of inverse of the sample covariance matrix is not only computationally expensive but unstable. This method ~\cite{zbontar2021barlow} computes cross-correlation matrix and makes it close to identity matrix in Frobenius norm. This paper~\cite{vicreg} suggested a similar approach. Unlike the methods mentioned above, which compute the covariance matrix with the only positive pairs, singular value loss term in \ours computes the covariance matrix along the batch dimension (with the negative pairs) and helps global dispersion of embeddings by making the embeddings be distributed isotropically. To emphasize this aspect, we coinage our loss as singular value loss, which regularizes the consistency of singular values of the empirical covariance matrix.

\myparagraph{Multiview Loss.}
In supervised learning settings, batch repetition method~\cite{batch_repetition} is proposed to improve the image classification performance as well as training efficiency. Recent self-supervised learning based on contrastive learning usually uses two views of the same image as positive pairs. And it is trained to minimize the distance or maximize similarity of embeddings of those two views. Recent work~\cite{swav} suggested multi-crop method which maximizes the similarity between views more than 2. To reduce computational cost, it generates 2 views with high resolution ($224 \times 224$ for ImageNet) and several other views with low resolution ($96 \times 96$ for ImageNet). This method~\cite{whitening_for_self_supervised} generates multiple positive views to perform whitening among them. In contrast, our method uses multiple views of the positive views of the same images to compute the centroid and distance between embeddings and the centroid. We discuss the relationship between batch repetition method and multiview centroid loss in the appendix.

\myparagraph{Uniformity of Embeddings.}
Our work is aligned with~\cite{wang2020understanding} in that our method also seeks to distribute the embeddings as uniformly as possible on the embedding space. That work claims that the contrastive learning is to make embeddings be distributed uniformly on the hypersphere. Similarly, our work also tries to distribute the embeddings uniformly on the embedding space. That method reformulates contrastive loss as the sum of alignment loss and uniformity loss. The first term, alignment loss, aligns positive views. The second term, uniformity loss, makes the embedding distribution $\textit{uniform}$ on the surface of unit sphere. The uniformity loss is defined by Gaussian kernel. The difference to our method is that 1)~\cite{wang2020understanding} is based on constrastive method and 2)~\cite{wang2020understanding} hypothesizes the embedding space is hypersphere. However, our method seeks more general embedding space with the dynamical system modeling. The advantage of modeling dynamical system is that we can study
the motion of embeddings and control them with this model. We further study the effect of our loss terms to uniformity and alignment trade-off in depth in the appendix.

\section{Method} \label{sec:brownian_byol}
\subsection{BYOL Architecture}

We follow the recent BYOL architecture~\cite{grill2020bootstrap} that learns a joint embedding of an image $x \in \mathcal{X}$ with two networks -- consists of two neural networks referred to as the {\em online} (or fast learner) and {\em target} (or slow learner) network. For completeness, we summarize some of the key details of the BYOL architecture. As shown in Figure~\ref{fig:model_overviews}, the online network is trained to predict the target network's representation of the augmented view of the same image. This online network is parameterized by a set of learnable weights $\theta$ and consists of three consecutive components: a backbone $f_\theta$, a projection head $g_\theta$, and a prediction head $h_\theta$. The target network is parameterized by a set of weights $\xi$ and consists of two components: a backbone $f_\xi$ and a projection head $g_\xi$. The parameter $\xi$ is updated by the bias-corrected exponentially weighted moving average of the online network's parameter $\theta$ at each training step, i.e. $\xi_{t+1} = \tau_t \xi_t + (1-\tau_t) \theta_t$ where $\tau_t \in [0, 1]$ is a target decay rate. 

Two augmented views $v\triangleq t(x)$ and $v'\triangleq t'(x)$ are generated by applying image augmentations $t\sim \mathcal{T}$ and $t'\sim \mathcal{T'}$ given two distributions of image augmentations $\mathcal{T}$ and $\mathcal{T}'$. The online network outputs $z\triangleq g_\theta(f_\theta(v))$ from the first augmented view $v$, while the target network produces $z'\triangleq g_\xi(f_\xi(v'))$ from the second augmented view $v'$. A prediction from the online network $p\triangleq h_\theta(z)$ is then $l_2$-normalized to compute the cosine similarity loss $\mathcal{L}_{\textnormal{byol}}$ by measuring mean squared error between the normalized prediction $p$ and the normalized target predictions $z'$:
\begin{align}
    \mathcal{L}_{\text{byol}} (\theta, \xi; \mathcal{X}) := \norm{\hat{p}-\hat{z}'}^{2} = 2-2\frac{\langle p, z' \rangle}{||p||_{2} \cdot ||z'||_{2}}
    \label{eq:byol_loss}
\end{align}
where $\hat{p}={p}/{\norm{p}}$ and $\hat{z}'={z'}/{\norm{z'}}$. Note that the loss $\mathcal{L}_{\text{byol}}$ is optimized with respect to $\theta$ only, but not $\xi$. The gradient does not back-propagate through the target network as depicted by {\em stop-gradient} in Figure~\ref{fig:model_overviews}. After training, both the prediction head $h_\theta$ and the projection head $g_\theta$ are discarded and the representations $z$ of the online network are used for downstream tasks.

\subsection{\ours}
Built upon BYOL architecture, we use the following loss $\mathcal{L}(\theta, \xi; \mathcal{X})$ (instead of using $\mathcal{L}_{\text{byol}}$) that consists of the following three loss terms: (i) multiview centroid loss $\mathcal{L}_{c}$, (ii) singular value loss $\mathcal{L}_{s}$, and (iii) Brownian diffusion loss $\mathcal{L}_{b}$. The overall loss is defined as follows:
\begin{align}
\mathcal{L}(\theta, \xi; \mathcal{X}) &= \mathcal{L}_{c}(\theta, \xi; \mathcal{X})+\lambda_{s} \mathcal{L}_{s}(\theta; \mathcal{X})+\lambda_{b} \mathcal{L}_{b}(\theta; \mathcal{X})  \label{eq:our_total_loss}
\end{align}

\myparagraph{Multiview Centroid Loss.} 
As opposed to BYOL, we train the online network to predict the target network's {\em centroid} representation of differently augmented multi-views of the same image. Given an image $x\in\mathcal{X}$, we generate $K$ differently augmented views (i.e. multi-view): $v_j \triangleq t_j(x)$ and $v_l \triangleq t_l(x)$ for $j,~l\in\{1, 2, \dots, K\}$ by applying stochastic image augmentations $t_j, t_l \sim \mathcal{T}$. Given $K$ outputs from the target network, $z'_l=g_\xi(f_\xi(v'_l))$, we use the geometric center of these $K$ outputs as the centroid representation, i.e. $\frac{1}{K}\sum_{l=1}^K \hat{z}'_{l}$ where $\hat{z}'_l={z'_l}/{\norm{z'_l}}$. Lastly, we compute the sum of $L_2$ loss between the target network's centroid representation of embeddings of $K$ different views  $\{\hat{z}'_{l}\}_{1}^{K}$ and the online network's representation -- thus, this loss applies an attractive force to pull together multiple augmented representations of the same image (positive pairs) into the geometric centroid as the pivot to cluster embeddings.

Ultimately, we the following multiview centroid loss $\mathcal{L}_{c}$:
\begin{align}
    \mathcal{L}_{c}(\theta, \xi; \mathcal{X}) &= \frac{1}{K} \sum_{j=1}^K \norm{\hat{p}_j-\frac{1}{K} \sum_{l=1}^K \hat{z}'_{l}}^2 \label{eq:our_centroid_loss}
\end{align}
where $\hat{p}_j={p_j}/{\norm{p_j}}$ is $l_2$-normalized predictions from the online network for the augmented view of the same input, i.e. $p_j=h_\theta(g_\theta(f_\theta(v_j)))$. Note that, minimizing Eq. \ref{eq:our_centroid_loss} is mathematically identical to minimizing pairwise distance between $\{\hat{p}'_{j}\}_{1}^{K}$ and $\{\hat{z}'_{l}\}_{1}^{K}$. Therefore, this loss generates a stronger attractive force that aggregates the embeddings of the same image that BYOL loss in Eq. \ref{eq:byol_loss}. 

\myparagraph{Brownian Diffusion Loss.} 
We use a dispersive loss, called Brownian diffusion loss, that induces a Brownian motion (or a random walk) of the online network's representation $p_j$ of $j$-th augmented view of an input. A $d$-dimensional random vector $n\in\mathbb{R}^d$ is sampled from unit normal distribution, i.e. $n \sim \mathcal{N}(0, I_d)$ with an identity matrix $I_d$. Our Brownian diffusion loss is defined as follows:
\begin{align}
    \mathcal{L}_{b}(\theta; \mathcal{X}) &= \frac{1}{K} \sum_{j=1}^K \langle \hat{n}, \hat{p}_j \rangle \label{eq:our_brownian_loss}
\end{align}
where $\hat{n} = n/\norm{n}$. The noise vector $\hat{n}$ drives a diffusive motion by pushing particles in the embedding space in radial direction, which is uniformly sampled on the unit hyper-sphere.
 
Importantly, we use the same random vector $\hat{n}$ for the all augmented embeddings of the given image. This implies that the positive pairs which share the similar semantics are not spread apart. In contrast, the views from the different image moves to the different direction and the direction is likely to be orthogonal to other images' moving direction. I.e. Brownian diffusion loss disperses the embeddings locally, which gives implicit contrastive effect between embeddings of different images. 

We observe that our Brownian diffusion loss is critical to prevent mode-collapse~\cite{grill2020bootstrap}. As the target network's parameter is updated by the exponentially weighted moving average of the online network's parameter at each training step (given a high target decay rate), the change of the target network's representation is relatively slower than that of the online network (effectively $\frac{1}{1-\tau}$ times slower). Such an imbalance may cause a mode collapse as the online network's representation can quickly collapse into a single point without any repulsive force between them.

\myparagraph{Singular Value Loss.} 
Lastly, we use the singular value loss $\mathcal{L}_w$ that decorrelates the different feature dimensions of the projections $\hat{p}$ to prevent these dimensions from conveying the same information, thus avoid a dimension collapse. We minimize the following Euclidean distance between the empirical covariance matrix of the embeddings and the identity matrix $I_d$ -- thus, we penalize the off-diagonal coefficients of the covariance matrix and make the distribution ball-shaped. Let the $p_{ij}$ be i-th batch and j-th augmented embeddings. Then the empirical covariance matrix of j-th augmented embeddings $S_j$ is:
\begin{align}
    S_j = \frac{1}{n-1} \sum_{i=1}^n (p_{ij} - \bar{p}_j) (p_{ij} - \bar{p}_j)^T
\end{align}
where $n$ is the number of batches and $\bar{p_j} = \frac{1}{n} \sum_{i=1}^n p_{ij}$. Then we define singular value loss as:
\vspace{-2mm}
\begin{align}
    \mathcal{L}_{s}(\theta; \mathcal{X}) &= \frac{1}{K} \sum_{j=1}^K {\normfro{ S_j - I_d}^2} \\
    &= \frac{1}{K} \sum_{j=1}^K \sum_{i=1}^d (\sigma_{ij} - 1)^2
    \label{eq:whitening_loss}
\end{align}
where $\sigma_{ij}$ is singular values of the covariance matrix, $S_j$. We found that this loss improves when corporated with Brownian diffusion loss. 

Some prior works~\cite{vicreg,whitening_for_self_supervised,zbontar2021barlow} justify whitening loss as removing correlations between different embeddings. In our method however, we treat singular value loss as a dispersive force that encourages uniformity of the embedding distribution. Even though Brownian diffusion loss addresses local dispersion in the embedding space, singular value loss exerts the force to regularize the shape of embedding distribution to be globally spherical at large scale.

\section{Experiments} \label{sec:experiment}

\begin{table*}[h]
  \caption{Classification accuracy (top-$1$ and top-$5$) of a linear classifier and 5-nearest neighbors ($5$-NN) classifier for different loss functions on two visual benchmarks: ImageNet-100~\cite{deng2009imagenet} and STL-10~\cite{stl10}. Note that BYOL (1st row), W-MSE 4 (5th row), and ours (the bottom 2 rows) are based on a ResNet-18 encoder, while others on a more expressive ResNet-50 encoder. $^\dagger$: scores are from our reproduction.}
  \label{table:classification_performance}
  \centering
  \resizebox{.85\linewidth}{!}{%
  \begin{tabular}{lccccccc}
    \toprule 
    \multirow{2}{*}{Method} & \multirow{2}{*}{Backbone} & & \multicolumn{3}{c}{ImageNet-100~\cite{tian2020contrastive}} & \multicolumn{2}{c}{STL-10~\cite{stl10}}  \\\cmidrule{4-8}
    & & & Top-1 (\%) & Top-5 (\%) & 5-NN (\%) & Top-1 (\%) & 5-NN (\%) \\ \midrule
    BYOL$^\dagger$~\cite{grill2020bootstrap} & ResNet-18 & & 71.56 &  91.18 & 63.18 & 89.50 & 85.15  \\
    MoCo~\cite{moco_he_2020_cvpr} & ResNet-50 & & 72.80 & 91.64 & - & - & -  \\
    SimCLR~\cite{simclr} & ResNet-50 &  & - & - & - & 90.51 & 85.68  \\
    Wang and Isola~\cite{wang2020understanding} & ResNet-50 &  & 74.60 & 92.74 & - & - & -  \\
    W-MSE 4~\cite{whitening_for_self_supervised} & ResNet-18 &  & 79.02  & 94.46  & 71.32 & 91.75 & 88.59  \\ \midrule
    Ours $(K=4)$ & ResNet-18 & & 80.38 & 94.92 & 74.30 & 93.00 & 90.38  \\
    Ours $(K=8)$ & ResNet-18 & & {\bf{81.56}} & {\bf{95.20}} & {\bf{75.24}} & {\bf{93.19}} & {\bf{90.56}}  \\
    \bottomrule
  \end{tabular}}
\end{table*}

\begin{table}[h]
  \caption{Downstream task result comparison on ImageNet. The backbone architecture for all the methods is ResNet-50. Note that the baseline results are from~\cite{chen2020simsiam} and ~\cite{whitening_for_self_supervised}. Bold face is the best accuracy and the underline is the second best accuracy. $\dagger$ means that our model is trained 300 epochs.}
  \label{table:imagenet_classification_performance}
  \centering
  \begin{tabular}{lrccc}
    \toprule 
    \multirow{2}{*}{Method} & \multirow{2}{*}{Batch Size} & \multicolumn{3}{c}{Epoch}  \\
    \cmidrule{3-5}
    & & 100 & 200 & 400 \\
    \midrule
    BYOL~\cite{grill2020bootstrap} & 4,096 & 66.5 &  70.6 & 73.2  \\
    SimCLR~\cite{simclr}  & 4,096 & 66.5 & 68.3 & 69.8  \\
    MoCo-v2~\cite{moco_he_2020_cvpr} & 256 & 67.4 & 69.9 & 71.0   \\
    W-MSE 4~\cite{whitening_for_self_supervised} & 1,024 & 69.3 & - & 72.56  \\ 
    SwAV~\cite{swav} (w.o. multi-crop) & 4,096 & 66.5 & 69.1 & 70.7  \\
    SwAV~\cite{swav} (with multi-crop) & 4,096 & \textbf{72.1} & \textbf{73.9} & \textbf{74.6}  \\
    SimSiam~\cite{chen2020simsiam} & 256 & 68.1 & 70.0 & 70.8 \\\midrule
    Ours $(K=4)$  & 512 & $\underline{70.7}$ & $\underline{73.8}$ & $\textbf{74.6}^\dagger$	 \\
    \bottomrule
  \end{tabular}
  \vspace{-1em}
\end{table}

\begin{table}[h]
  \caption{Semi-Supervised classification result comparison on the subsets of ImageNet. We finetune the classifier and the encoder with $1\%$ and $10\%$ of labeled data of ImageNet. We report top-1 and top-5 accuracy. Bold face is the best accuracy.}
  \label{table:semi_supervised_performance}
  \centering
  \begin{tabular}{lcccc}
    \toprule 
    \multirow{2}{*}{Method} & \multicolumn{2}{c}{Top-1 (\%)} & \multicolumn{2}{c}{Top-5 (\%)}  \\
    \cmidrule{2-5}
    & $1\%$ & $10\%$ & $1\%$ & $10\%$ \\
    \midrule
    SimCLR~\cite{simclr}  & 48.3 & 65.6 & 75.5 & 87.8  \\
    BYOL~\cite{grill2020bootstrap} & 53.2 & 68.8 &  78.4 & 89.0  \\
    VICReg~\cite{vicreg} & 54.8 & 69.5 & 79.4 & 89.5   \\
    SwAV~\cite{swav} (with multi-crop) & 53.9 & 70.2 & 78.5 & 89.9  \\
    Barlow Twins~\cite{zbontar2021barlow} & 55.5 & 69.7 & 79.2 & 89.3 \\
    OBoW~\cite{OBoW} & - & - & \textbf{82.9} & \textbf{90.7} \\
    \midrule
    Ours $(K=4)$  & \textbf{58.6} & \textbf{70.6} & 81.9 & 90.1	 \\
    \bottomrule
  \end{tabular}
\end{table}

\begin{table}[h]
  \caption{kNN classification result comparison on ImageNet. We report accuracy with 20-NN and 200-NN. }
  \label{table:knn_performance}
  \centering
  \begin{tabular}{lrcc}
    \toprule 
    Method & 20-NN (\%) & 200-NN (\%)  \\
    \midrule
    NPID~\cite{npid} & - & 46.5 \\
    LA~\cite{la} & - & 49.4 \\
    PCL~\cite{pcl} & 54.5 & - \\
    VICReg~\cite{vicreg} & 64.5 & 62.8  \\
    SwAV~\cite{swav} (with multi-crop) & 65.7 & 62.7 \\
    \midrule
    Ours $(K=4)$  & \textbf{66.2} & \textbf{63.0}	 \\
    \bottomrule
  \end{tabular}
\end{table} 

\subsection{Evaluation of Representations with \ours} \label{sec:eval_representation_with_bmwreg}
\myparagraph{Evaluation on ImageNet-100 and STL-10.}
Following the linear evaluation protocol, we train a simple linear classifier with the frozen representations from our encoder, which is pre-trained with our \ours. We first evaluate the performance of the encoder on a small-size ImageNet-100~\cite{tian2020contrastive} and a mid-size STL-10 datasets. We observe in Table~\ref{table:classification_performance} that the performance of \ours generally outperforms other state-of-the-art approaches on both datasets, especially we observe a large gain on the STL-10 dataset. The performance gain is more apparent that the following three approaches, MoCo, SimCLR, and Wang and Isola, use a more expressive ResNet-50-based backbone than our ResNet-18-based backbone. We also observe that the quality of the learned representation improves as the number of views $K$ increases (compare bottom two rows). 

\myparagraph{Evaluation on ImageNet.}
We further evaluate the representations obtained after self-supervised pre-training with \ours on the large-scale ImageNet dataset with two evaluation metrics: 1) linear evaluation protocol and 2) semi-supervised learning with the subsets of ImageNet and 3) kNN classification. For linear evaluation protocol, likewise to Table~\ref{table:classification_performance}, dashed-line means that the original paper did not report the corresponding value. We observe in Table~\ref{table:imagenet_classification_performance} that the performance of \ours outperforms other approaches and gets the result compatible to SwAV with multi-crop, which may confirm that the effectiveness of \ours for learning the better visual representation. Especially, compared to other baselines which are trained for 400 epochs, our method is only trained for 300 epochs. This implies that it has enough room to improve a lot. Again, note that ours uses a smaller batch size than alternatives except MoCo-v2 (i.e. 512 vs. 4096 or 1024), but shows the matched or better performance. 

To evaluate semi-supervised leaning ability of our method, we report top-1 and top-5 accuracy over $1\%$ and $10\%$ of ImageNet subsets. The experiment results are in Table~\ref{table:semi_supervised_performance}. For both $1\%$ and $10\%$ subsets, our method outperforms baselines, when we compare methods with top-1 accuracy. Especially, in the fine-tuning result with $1\%$ subset of ImageNet dataset (see 1st column in Table~\ref{table:semi_supervised_performance}), our method surpasses with the large margin. For top-5 accuracy, our method gets matched performance with ~\cite{OBoW} and outperforms other methods. 

kNN evaluation results are in Table~\ref{table:knn_performance}. We report the accuracy of 20-NN and 200-NN classification results. Our method outperforms baselines.  

\begin{table}[h]
  \caption{Evaluation of the representations pretrained with \ours on various downstream tasks: 1) the performance linear classifier on top of frozen ResNet-50 backbone and 2) object detection with fine-tuning. For the linear probing, we report mAP for VOC07~\cite{pascal_voc} benchmark, Top-1 accuracy ($\%$) for Places~\cite{places_dataset} and iNaturalist2018~\cite{inat2018} benchmarks. For the object detection task, we report $\text{AP}^{50}$, $\text{AP}^{75}$, and $\text{AP}^{\text{all}}$ for VOC07+12 benchmark.}
  \label{table:transfer_learning_comparison}
  \centering
  \begin{tabular}{lccccccccccc}
    \toprule 
    \multirow{2}{*}{Method} & &\multicolumn{3}{c}{Classification (\%)} & & & \multicolumn{3}{c}{VOC Detection} \\\cmidrule{2-10}
    & & VOC07 & Places & iNat18 & & & $\text{AP}^{50}$ & $\text{AP}^{75}$ & $\text{AP}^{\text{all}}$\\ \midrule
    BoWNet~\cite{bownet} & & 79.3 & 51.1 & - & & & 81.3 & 61.1 & 53.5 \\
    MoCo v2~\cite{moco_he_2020_cvpr} & &  86.4 & 51.8 & 38.6 & & & 82.4 & 63.6 & 57.0  \\
    PIRL~\cite{pirl} & & 81.1 & 49.8 &  - & & & 80.7 & 59.7 & 54.0 \\
    OBoW~\cite{OBoW} & & 89.3 &  56.8 & - & & & 82.9 & 64.8 & 57.9 \\
    BYOL~\cite{grill2020bootstrap} & &  86.6 & 54.0 & 47.6 & & & 81.4 & 61.1 & 55.3 \\
    SimSiam~\cite{chen2020simsiam} & & - & - & - & & & 82.4 & 63.7 & 57.0 \\
    Barlow Twins ~\cite{zbontar2021barlow} & & 86.2 & 54.1 & 46.5 & & &  82.6 & 63.4 & 56.8  \\
    SwAV~\cite{swav} & & 88.9 & 56.5 & 48.6 & & & 82.6 & 62.7 & 56.1 \\
    PixPro~\cite{pixpro} & & - & - & - & & & 83.8 & 67.7 & 60.2  \\
    \midrule
    Ours $(K=4)$ & & 87.8 & 56.4 & 47.9 & & & 83.0 & 63.5 & 56.7 \\ 
    \bottomrule
  \end{tabular}
  \vspace{-2em}
\end{table}

\begin{table}[h]
  \caption{Comparison of the quality of representations between BYOL~\cite{grill2020bootstrap} and ours on the STL-10 dataset~\cite{stl10}. The Top-1 classification accuracy is reported with different types of normalization techniques: a batch normalization (BN)~\cite{batch_norm} and a layer norm (LN)~\cite{layer_norm}. To see the effect of our proposed Brownian Diffusive Loss, $\mathcal{L}_b$, we also report scores of BYOL with $\mathcal{L}_b$ (4th row).}
  \label{table:repulsive_coeff_effect_layernorm}
  \centering
  \begin{tabular}{lcccc}
    \toprule 
    Method & Norm. Layer & Batch Size & $\lambda_b$ & Top-1 (\%)\\ \midrule
    BYOL & BN & 256 & 0 & 89.5 \\
    Ours & BN & 256 & $5\times 10^{-2}$ & 91.4 \\\midrule
    BYOL & LN & 256 & 0 & 10.6 \\
    BYOL + our $\mathcal{L}_b$ & LN & 256 & $5\times 10^{-3}$ & 75.3\\ 
    BYOL & LN & 1024 & 0 & 10.6 \\\midrule
    Ours & LN & 256 & $5\times 10^{-4}$ & 80.7 \\
    Ours & LN & 256 & $5\times 10^{-3}$ & 82.3 \\
    Ours & LN & 256 & $5\times 10^{-2}$ & 78.7 \\
    \bottomrule
  \end{tabular}
  \vspace{-2em}
\end{table}
\subsection{Transfer Learning on Various Downstream Tasks}
We further evaluate the transferability of the features trained with \ours on ImageNet via transferring the features to various downstream tasks. In Table~\ref{table:transfer_learning_comparison}, we compare the performance of \ours with baselines. We first report the linear classification result on VOC07~\cite{pascal_voc}, Places205~\cite{places_dataset} and iNaturalist~\cite{inat2018} visual benchmarks. Each of benchmarks is to evaluate 1) multi-label classification 2) scenic scenario and 3) fine-grained classifcation. We evalute the performance of linear classifier on top of the frozen ResNet-50 encoder pretrained with \ours method. We report mAP for VOC07 dataset and top-1 accuracy ($\%$) for other benchmarks. We observe that our method shows generally matched results compared with alternatives. 

For object detection task, we finetune pre-trained ResNet-50 backbone with the PASCAL VOC07+12 object detection benchmark~\cite{pascal_voc}. We use Faster R-CNN~\cite{fasterrcnn} with C4 backbone as our baseline model. We report $\text{AP}^{50}$, $\text{AP}^{75}$, and $\text{AP}^{\text{all}}$. We observe in Table~\ref{table:transfer_learning_comparison} that our model shows generally matched results compared with alternatives except for PixPro~\cite{pixpro}, which is proposed for . For $\text{AP}_{50}$, our method performs better than the baselines, while our method shows matched or slightly lower performance than other approaches. 

We report instance segmentation result on COCO dataset in the appendix.

\subsection{Brownian Diffusive Loss against Mode Collapse} \label{sec:repuslive_force_effect}
BYOL~\cite{grill2020bootstrap} successfully uses only pairs of positives, but the reason why the online and target networks can avoid a so-called mode collapse, i.e. representations of all the examples are mapped to the same point in the embedding space, is not yet clearly explained. Existing work~\cite{untitled-ai,chen2020simsiam,tian2020contrastive,richemond2020byol_wo_bn} discussed that the use of the Batch Norm (BN) implicitly contributes to avoiding generating a collapsed representation. Especially, the original authors of~\cite{grill2020bootstrap} show that BYOL works without BN~\cite{richemond2020byol_wo_bn}. However, those methods are impractical in terms of restricting the network architecture design and this fact implies that these approaches are suboptimal. In our work, we propose to use Brownian diffusive loss, $\mathcal{L}_b$, which pushes embeddings into the radial direction to be uniformly sampled on the unit hyper-sphere. This helps to avoid collapsed representations without the need of using the Batch Norm (BN). We further discuss this in the appendix.

In Table~\ref{table:repulsive_coeff_effect_layernorm}, we empirically observe that BYOL suffers from a mode collapse when we replace the Batch Norm (in the prediction and projection heads) with another normalization technique, a Layer Norm (compare 1st vs. 3rd row). The top-1 classification accuracy is largely degraded from 89.5\% to 10.6\%, i.e. mode collapsed. Ours with the Brownian diffusive loss $\mathcal{L}_b$ was not the case (compare 2nd vs. 6th row). Though we observe a slight degradation in the top-1 classification accuracy, ours sufficiently avoid collapsed representations. Further, we evaluate the BYOL with our Brownian diffusive loss to demonstrate its effectiveness against a mode collapse. We observe that our Brownian diffusive loss helps avoid collapsed representations (compare 3rd vs. 4th rows). We also observe that the quality of representations depends on the strength of the hyperparameter $\lambda_b$ where we obtain the best performance with $\lambda_b=5\times 10^{-4}$. We observe a tension as we see a smaller or larger $\lambda_b$ slightly degrades the quality of representations.

\subsection{Comparison with Multi-Crop Method}
We further compare Multiview centroid loss with the multi-crop method. The main difference between multiview centroid loss and multi-crop in SwAV is that our method uses the same resolution across all views while multi-crop uses low resolutions. We observe in Table~\ref{table:comparison_multicrop_multiview} that a BYOL model with the multi-crop method shows a degradation (compare 1st vs. 2nd row), while \ours improves the performance of BYOL with a large margin (compare 1st vs. 3rd). This fact is also reported in~\cite{vicreg}.

Here, we describe the details of experiment. For a fair comparison, we implement the multi-crop method in the BYOL framework. The multi-crop method generates 2 views with full resolution ($224 \times 224$ for ImageNet) and $V$ views with low resolution ($96 \times 96$ for ImageNet). Cropping small parts of an image is used to generate low-resolution images. We choose $V = 6$ following~\cite{swav}. To apply the multi-crop method to BYOL, we reformulate BYOL loss (i.e. Eq.~\ref{eq:byol_loss}) as follows:

\begin{align*}
    \mathcal{L}_{\text{mc-byol}} (\theta, \xi; \mathcal{X}) &:= \sum _{i, j}^{V+2}\norm{\hat{p_i}-\hat{z_j}'}^{2} \mathbbm{1}(i\neq j) \\
    &= \sum _{i, j}^{V+2} \bigg ( 2-2\frac{\langle p_i, z_j' \rangle}{||p_i||_{2} \cdot ||z_j'||_{2}} \bigg ) \mathbbm{1}(i\neq j)
    \label{eq:mc_byol_loss}
\end{align*}
which shows that the multi-crop method minimizes the distance between pairs of embeddings, while our Multiview centroid loss minimizes the distance between each view and the geometric centroid of multi-views.

\begin{table}[h]
  \caption{Comparison accuracy of downstream image classification task on ImageNet between multiview loss and multi-crop~\cite{swav}. We apply multi-crop method to BYOL. }
  \label{table:comparison_multicrop_multiview}
  \centering
  \begin{tabular}{lcccc}
    \toprule 
    Method & 100 epochs & 200 epochs & 300 epochs \\ \midrule
    BYOL~\cite{grill2020bootstrap} & 65.9 & 70.1 & 72.3 \\
    BYOL+multi-crop & 65.8 & 68.7 & 70.3 \\
    Ours ($K=4$) & 70.2 & 73.6 & 74.4 \\
    \bottomrule
  \end{tabular}
  \vspace{-2em}
\end{table}
\subsection{Ablation Studies}
Table~\ref{table:ablation_studies} shows our ablation study to see the effect of our proposed three regularizations: (1) Multiview centroid loss $\mathcal{L}_c$, (2) Brownian diffusive loss $\mathcal{L}_b$, and (3) Singular value loss $\mathcal{L}_s$. Given the BYOL model as a baseline, we apply different combinations of our regularizations and measure the quality of representations following the linear evaluation protocol. We report scores on the ImageNet-100 dataset. We use \cmark~and \xmark~to indicate {\it{with}} and {\it{without}}, respectively. Note that we set $\lambda_b$ and $\lambda_w$ by default as 0.5 and $4.0\times 10^{-3}$, respectively. 

We first observe that a significant performance gain is obtained with our Multiview centroid loss $\mathcal{L}_c$ (compare 1st vs. 5th and 9th). The quality of the learned representations consistently improves as the number of views $K$ increases. Since BYOL uses 2 views ($K=2$) for training, doubling the number of views provides more than $6\%$ performance gain. The other two regularizations, Brownian diffusive loss $\mathcal{L}_b$ and Singular value loss $\mathcal{L}_s$, also consistently improve the overall classification accuracy. For example, the classification performance improves $0.92\%$ with the Brownian diffusive loss (compare 1st vs. 3rd) and the Singular value loss (compare 1st vs. 2nd). Such performance gain becomes more apparent with the Multiview centroid loss where we obtain a larger gain: $1.44\%$ with the Singular value loss and $1.5\%$ with the Brownian diffusive loss. Concretely, applying all our proposed regularizations together shows the best performance. 

We further study the sensitivity of the tuning of the two new hyperparameters $\lambda_s$ and $\lambda_b$. We report the result in the supplementary.

\begin{table}[t]
  \caption{Ablation study to study the effect of our proposed three regularizations: (1) Multiview centroid loss $\mathcal{L}_c$, (2) Brownian diffusive loss $\mathcal{L}_b$, and (3) singular value loss $\mathcal{L}_w$. Note that we compare the top-1 classification accuracy (in \%) of a linear classifier on the ImageNet-100 dataset.}
  \label{table:ablation_studies}
  \centering
  \begin{tabular}{ccccc}
    \toprule 
    Method & $\mathcal{L}_c$ & $\mathcal{L}_b$ & $\mathcal{L}_s$ & Acc. (\%) \\ \midrule
    BYOL & \xmark & \xmark & \xmark & 71.92 \\
    BYOL & \xmark & \xmark & \cmark & 72.84 \\
    BYOL & \xmark & \cmark & \xmark & 72.84 \\
    BYOL & \xmark & \cmark & \cmark & 72.41 \\ \midrule
    Ours ($K=4$) & \cmark & \xmark & \xmark & 78.24 \\
    Ours ($K=4$) & \cmark & \xmark & \cmark & 79.68 \\
    Ours ($K=4$) & \cmark & \cmark & \xmark & 79.74 \\
    Ours ($K=4$) & \cmark & \cmark & \cmark & 80.38 \\ \midrule
    Ours ($K=8$) & \cmark & \xmark & \xmark & 79.54 \\
    Ours ($K=8$) & \cmark & \xmark & \cmark & 79.96 \\
    Ours ($K=8$) & \cmark & \cmark & \xmark & 80.28 \\
    Ours ($K=8$) & \cmark & \cmark & \cmark & 81.56 \\
    \bottomrule
  \end{tabular}
  \vspace{-1em}
\end{table}

\section{Conclusion}     
In this work, we have explored multiview, singular value regularization and Brownian diffusion methods for self-supervised learning. Each method implicitly induces contrastive effect, which stabilizes the the training of self-supervised learning. Our method achieves a good downstream task performance for instance classification as well as various transfer learning such as object detection, semantic segmentation. 


\section{Appendix}


\section*{Content}
This supplementary material provides implementation details (Section~\ref{sec:implementation_details}) including training strategy, architectures, and image augmentations. We also provide evaluation details (Section~\ref{sec:evaluation_details}) including linear evaluation protocol, semi-supervised learning setting, k-NN classification, and transfer learning on various downstream tasks. We also report supplemental experimental results of instance segmentation on COCO dataset (Section~\ref{sec:additional_results}). In addition, this supplementary presents ablation study results. Lastly, we compare our method with ~\cite{batch_repetition} and ~\cite{wang2020understanding} in detail.

\section{Implementation Details} \label{sec:implementation_details}
We first provide implementation details of our method. We would emphasize that our code will be made publicly available upon publication. In Section~\ref{sec:training_strategy} and \label{sec:architecture}, we explain details of our training strategy and architectures. Next, in Section~\ref{sec:image_augmentations}, we explain details of the stochastic image data augmentation used in our experiment. 

    
    
    
    

    
    
    
    
    
    
    
    
    

\subsection{Training Strategy.} \label{sec:training_strategy}
We utilize the Layer-wise Adaptive Rate Scaling (LARS)~\cite{lars} optimizer that is known to effectively overcome large-batch training difficulties. We also use the learning rate scheduler that applies a cosine decay function~\cite{loshchilov2017cosine_lr_decay} without restarts to an optimizer step. As suggested by \cite{goyal2017accurate}, we apply a learning rate warm-up for the first 10 epochs where we start training with a small safe learning rate, which is slowly increased to the max learning rate linearly. The max learning rate is $\texttt{base\_lr} \times \frac{\texttt{batch size}}{256} \times K$ ~\cite{goyal2017accurate}. We set the base learning rate to 0.4 for ImageNet-100, 0.5 for STL-10 dataset and 0.15 for ImageNet datasets. Unless otherwise stated, we set the batch size to 512. The weight decay parameter is set to $1\times10^{-5}$. We exclude biases and parameters in batch normalization layer following BYOL~\cite{grill2020bootstrap}. We train the model for 320 epochs for ImageNet-100 and STL-10 benchmarks and 300 epochs for ImageNet with 8 V100
16GB GPUs.

\subsection{Architectures} \label{sec:architecture}
For a fair comparison, we use ResNet-18~\cite{he2015resnet} as a backbone network architecture for STL-10 and ImageNet-100 datasets and ResNet-50 as a backbone for ImageNet dataset, which are widely experimented with conventional approaches for the self-supervised representation learning task. Following BYOL~\cite{grill2020bootstrap}, the projection heads (i.e. $f_\theta$ and $g_\xi$ in Figure 2 in the main paper) and the prediction head of the online network (i.e. $h_\theta$) use a 2-layer fully connected network with ReLU~\cite{relu} as an activation function. We tune the size of hidden layers and output layers of projection and prediction heads, when the backbone network is ResNet-18. We use $512$ hidden layer size and $128$ output layer size instead of $2048$ hidden units and $256$ output size, which are used in BYOL. We apply batch normalization layer~\cite{batch_norm}. Also, we experiment various normalization layers including weight standardization~\cite{weight_standardization} and layer normalization~\cite{layer_norm} to show that our method does not suffer from mode collapse without batch normalization.

\begin{table}
  \caption{Image augmentation parameters}
  \label{table:image_augmentation_parameters}
  \centering
  \begin{tabular}{lc}
    \toprule 
    Image Augmentation Parameters & Values  \\
    \midrule
    1. Random Crop Probability & $1.0$ \\
    2. Flip Probability & $0.5$ \\
    3. Color Jittering Probability & $0.8$ \\
    4. Brightness Adjustment Max Intensity & $0.4$ \\
    5. Contrast Adjustment Max Intensity & $0.4$ \\
    6. Saturation Adjustment Max Intensity & $0.2$ \\
    7. Hue Adjustment Max Intensity & $0.1$ \\
    8. Color Dropping Probability & $0.2$ \\
    9. Gaussian Blurring Probability & $1.0$ \\
    10. Solarization Probability & $0.2$ \\
    \bottomrule
  \end{tabular}
\end{table}

\subsection{Image Augmentations} \label{sec:image_augmentations}
We use a stochastic data augmentation operator that is sampled from the family of augmentations $\mathcal{T}$ and results in a randomly augmented view of any given data example. Following SimCLR~\cite{simclr}, our data augmentation module sequentially applies the following four augmentations: (1) random cropping followed by resizing back to the original size, (2) aspect-ratio changes, (3) random flipping in the horizontal direction, (4) random color distortion (i.e. jitter and lighting). Detailed augmentation parameters are in Table~\ref{table:image_augmentation_parameters}.

\section{Evaluation Details} \label{sec:evaluation_details}
In this section, we provide relevant information for evaluation of our method. 
\subsection{Linear Evaluation Protocol} \label{sec:linear_evaluation_protocol}
We use the linear evaluation protocol~\cite{kolesnikov2019revisiting}, which is the standard practice to evaluate the quality of the learned image representations. Using the trained encoder as the feature extractor, we train a linear classifier as a post-hoc manner, i.e. a simple image classifier given a set of features. Then, we measure its classification accuracy on the test set as a proxy of the quality of the learned representations. Note that the encoder is frozen during the evaluation phase. We use the following three standard image classification benchmarks: (1) STL-10~\cite{stl10}, (2)     ImageNet-100~\cite{tian2020contrastive}, and (3) ImageNet~\cite{deng2009imagenet}. Note that ImageNet-100 contains only 100-class examples that are randomly sampled from ImageNet 
\subsection{Semi-Supervised Learning} \label{sec:semi_supervised_learning_protocol}
We also evaluate the semi-supervised learning ability of our method with subset of ImageNet training set. We finetune ResNet-50 encoder pretrained with our algorithm and the classifier on top of the encoder using $1\%$ and $10\%$ of ImageNet. These ImageNet splits can be found from the official implementation of ~\cite{simclr}. 
We mainly follow the semi-supervised learning protocol of ~\cite{grill2020bootstrap}. We use SGD with momentum of 0.9 and Nesterov, batch size of 1024. We use the separate learning rates for the classifier and the encoder. For fine-tuning task with $1\%$ ImageNet subset, we set learning rate of the classifier 2.0 and freeze the encoder. For fine-tuning task with $10\%$ ImageNet subset, we use the 0.25 as the learning rate of the classifier and $2.5\times 10^{-4}$ as the learning rate of the encoder. 

\subsection{k-NN Classification} \label{sec:k_nn_classification_protocol}
We closely follow the existing work~\cite{memory_bank,la} to evaluate the quality of representations learned by our model. We first collect representations from training and validation images with the frozen encoder. Then, we compute the classification accuracy of 20/200-nearest neighbor classifier. 


\subsection{Transferring to Downstream Tasks} \label{sec:transferring_learning_protocol}
To test the transferability of representations trained with our method on ImageNet, we perform transfer learning to various datasets: Places205, iNaturalist2018, Pascal VOC, and COCO.

\myparagraph{Image classification.} We train the linear classifier layer on top of the frozen ResNet-50 backbone pretrained with \ours. For VOC 07, we train a linear support vector machine (SVM). For other image classification benchmarks, iNaturalist 2018 and Places 205, we train the linear classifier with SGD with momentum of 0.9 and weight decay of $10^{-4}$. The batch size is 256 and learning rate is 0.2 and we reduce the learning rate by factor of 10 two times with equally spaced intervals. For Places205, the training epoch is 28 and for iNaturalist 2018, the training epoch is 84.

\myparagraph{Object detection.} Following previous works~\cite{Scaling_Benchmarking_SSL,swav,vicreg,OBoW}, we finetune the network on VOC07+12~\cite{pascal_voc} dataset using Faster-RCNN~\cite{fasterrcnn}. We report three metrics of the object detection, $\text{AP}_{\text{all}}$, $\text{AP}_{\text{75}}$ and $\text{AP}_{\text{50}}$. We use Detectron2~\cite{detectron2} to transfer our model to the object detection task. We set the initial learning rate 0.02. Other hyperparameters such as learning rate scheduling, warm-up steps are exactly same as~\cite{moco_he_2020_cvpr}.

\myparagraph{Instance Segmentation.} For instance segmentation task, we evaluate our model with COCO dataset. We closely follow~\cite{moco_he_2020_cvpr,zbontar2021barlow,vicreg}. We use Mask R-CNN FPN backbone. The backbone is initialized with our pretrained ResNet-50 backbone. We train the network for 90K iterations with a batch size of
16. A learning rate is 0.05 and reduced by a factor of 10 after 60K and 80K iterations. We linearly warm up the learning rate for 50 iterations.

\section{Results on COCO Instance Segmentation} \label{sec:additional_results}
We also evaluate the learned representation on COCO insstance segmentation task. We observe in Table~\ref{table:instance_segmentation} that our method shows competitive performance with other methods. Our method is better than BYOL~\cite{grill2020bootstrap} (3rd row), which is our main baseline. SwAV~\cite{swav} (5th row) shows similar performance to ours. Note that this method uses more augmentations than ours. 

\begin{table}
  \caption{Performance comparison for transfer learning on instance segmentation task on COCO dataset. We use \texttt{train2017} as training data and report the box detection AP ($\text{AP}^{\text{bb}}$) and instance segmentation AP ($\text{AP}^{\text{mk}}$) scores on \texttt{val2017} dataset.}
  \label{table:instance_segmentation}
  \centering
  \begin{tabular}{lrccc}
    \toprule 
    Method & $\text{AP}^{\text{bb}}$ & $\text{AP}^{\text{mk}}$ \\
    \midrule
    SimCLR~\cite{bownet} & 39.7 & 35.8 \\
    MoCo~\cite{moco_he_2020_cvpr} & 40.4 & 36.4 \\
    BYOL~\cite{grill2020bootstrap} & 41.6 & 37.2\\
    VICReg~\cite{vicreg} & 39.4 & 36.4\\
    SwAV~\cite{swav} & 41.6 & 37.8\\
    BarlowTwins~\cite{zbontar2021barlow} & 40.0 & 36.7 \\
    OBoW~\cite{OBoW} & 40.8 & 36.4\\
    \midrule
    Ours $(K=4)$ & 41.8 & 37.8 \\ 
    \bottomrule
  \end{tabular}
  \vspace{1em}
\end{table}

\section{Ablation Studies}
We perform ablation experiments to study the trade off between major hyperparameters in \ours, $\lambda_s$ and $\lambda_b$. In table~\ref{table:ablation_studies}, our experiment reports the top-1 classification accuracy on ImageNet-100. We train ResNet-18 with~\ours for 300 epochs with various combinations of $K \in \{2, 4, 8\}$, $\lambda_s \in \{0, 0.002, 0.004, 0.01\}$, and $\lambda_b \in \{0, 0.25, 0.5, 1.0, 2.0\}$. Note that the case of $K=2$ is the same as BYOL setting. Then, we train the linear classifier on top of frozen ResNet-18 backbone pretrained with \ours. Our study shows that the classification accuracy increases until $\lambda_s$=0.004$, \lambda_b=0.5$ for the cases of $K=4$ and $K=8$. Interestingly, both singular value loss and Brownian loss improve the performance for the case of BYOL ($K=2$).

\begin{table}
  \caption{Ablation studies to investigate the trade-off between losses in~\ours.}
  \label{table:ablation_studies}
  \centering
  \begin{tabular}{cccc|cccc|cccc}
    \toprule 
    $K$ & $\lambda_s$ & $\lambda_b$ & Acc. ($\%$) & $K$ & $\lambda_s$ & $\lambda_b$ & Acc. ($\%$) & $K$ & $\lambda_s$ & $\lambda_b$ & Acc. ($\%$)\\
    \midrule
    2 & 0 & 0 & 71.9 & 4 & 0 & 0 & 78.2 & 8 & 0 & 0 & 79.5 \\
    2 & 0 & 0.25 & 72.3 & 4 & 0 & 0.25 & 79.1 & 8 & 0 & 0.25 & 80.1 \\
    2 & 0 & 0.5 & 72.8 & 4 & 0 & 0.5 & 79.7 & 8 & 0 & 0.5 & 80.3 \\
    2 & 0 & 1.0 & 72.3 & 4 & 0 & 1.0 & 79.2 & 8 & 0 & 1.0 & 80.3 \\
    2 & 0 & 2.0 & 71.9 & 4 & 0 & 2.0 & 79.0 & 8 & 0 & 2.0 & 80.2 \\ \hline
    2 & 0.002 & 0 & 72.2 & 4 & 0.002 & 0 & 78.9 & 8 & 0.002 & 0 & 79.8 \\
    2 & 0.002 & 0.25 & 72.4 & 4 & 0.002 & 0.25 & 78.8 & 8 & 0.002 & 0.25 & 80.2\\
    2 & 0.002 & 0.5 & 72.4 & 4 & 0.002 & 0.5 & 79.1 & 8 & 0.002 & 0.5 & 80.9 \\
    2 & 0.002 & 1.0 & 72.1 & 4 & 0.002 & 1.0 & 79.2 & 8 & 0.002 & 1.0 & 81.1 \\
    2 & 0.002 & 2.0 & 71.9 & 4 & 0.002 & 2.0 & 79.2 & 8 & 0.002 & 2.0 & 80.8 \\ \hline
    2 & 0.004 & 0 & 72.8 & 4 & 0.004 & 0 & 79.7 & 8 & 0.004 & 0 & 80.0 \\
    2 & 0.004 & 0.25 & 72.8 & 4 & 0.004 & 0.25 & 80.2 & 8 & 0.004 & 0.25 & 80.9 \\
    2 & 0.004 & 0.5 & 72.4 & 4 & 0.004 & 0.5 & 80.4 & 8 & 0.004 & 0.5 & 81.6 \\
    2 & 0.004 & 1.0 & 72.1 & 4 & 0.004 & 1.0 & 80.1 & 8 & 0.004 & 1.0 & 81.5 \\
    2 & 0.004 & 2.0 & 71.2 & 4 & 0.004 & 2.0 & 79.9 & 8 & 0.004 & 2.0 & 81.3 \\ \hline
    2 & 0.01 & 0 & 71.1 & 4 & 0.01 & 0 & 79.3 & 8 & 0.01 & 0 & 79.9\\
    2 & 0.01 & 0.25 & 71.0 & 4 & 0.01 & 0.25 & 79.4 & 8 & 0.01 & 0.25 & 81.2 \\
    2 & 0.01 & 0.5 & 71.0 & 4 & 0.01 & 0.5 & 79.2 & 8 & 0.01 & 0.5 & 81.4 \\
    2 & 0.01 & 1.0 & 70.7 & 4 & 0.01 & 1.0 & 79.2 & 8 & 0.01 & 1.0 & 81.1 \\
    2 & 0.01 & 2.0 & 70.3 & 4 & 0.01 & 2.0 & 79.0 & 8 & 0.01 & 2.0 & 79.8 \\
    \bottomrule
  \end{tabular}
  \vspace{1em}
\end{table}

\section{Related Works}
In this section, we supplement Section 2. We compare our work with batch repetition method~\cite{batch_repetition}, uniformity loss~\cite{wang2020understanding}, and BYOL without BN~\cite{richemond2020byol_wo_bn} in detail.

\myparagraph{Batch Repetition.} In the Section 2, we mention batch repetition method~\cite{batch_repetition}. Similar to this method, our multiview centroid loss partially benefits from the fact that simply seeing the same image with different augmentations at each iteration, stabilizes and accelerates training in self-supervised settings. However, the main difference between~\cite{batch_repetition} and multiview centroid loss, is that multiview centroid loss considers the interactions between embeddings of the positive pairs.

\myparagraph{Uniformity of Embeddings.} In this section, we report uniformity score of~\ours and other baselines in Table~\ref{table:alignment_uniformity}. We train BYOL, BYOL with uniformity loss, BYOL+$\mathcal{L}_b+\mathcal{L}_s$ and \ours with ImageNet-100 with ResNet-18 backbone. Then, we evaluate each model with three metrics: 1) linear classifier accuracy 2) alignment loss and 3) uniformity loss. Here, both alignment loss and uniformity loss are introduced in~\cite{wang2020understanding}. Alignment loss, $\mathcal{L}_{\text{align}}$ is defined as mean squared error between positive pairs and uniformity loss, $\mathcal{L}_{\text{uniform}}$, is defined as the logarithm of the average
pairwise Gaussian potential between negative pairs. In Table~\ref{table:alignment_uniformity}, uniformity loss improves the performance of BYOL (1st row vs 2nd row), by decreasing uniformity loss. Ours shows lower uniformity loss, higher alignment loss and the better performance than other baselines. This strengthen the argument of~\cite{wang2020understanding} and ours, which argues that the optimal distribution trained with self-supervised method is uniformly on the embedding manifold.

\begin{table}
  \caption{Evaluating methods with $\mathcal{L}_{\text{align}}$ and $\mathcal{L}_{\text{uniform}}$}
  \label{table:alignment_uniformity}
  \centering
  \begin{tabular}{lccc}
    \toprule 
    Method & Acc.($\%$) & Alignment & Uniformity \\
    \midrule
    BYOL & 71.9 & 0.25 & -1.52 \\ 
    BYOL+$\mathcal{L}_{\text{uniform}}$ & 72.1 & 0.27 & -2.95 \\ 
    BYOL+$\mathcal{L}_b+\mathcal{L}_s$ & 72.8 & 0.26 & -2.92 \\ \midrule
    Ours ($K=4$) & 80.4 & 0.36 & -3.8 \\
    \bottomrule
  \end{tabular}
  \vspace{1em}
\end{table}

\myparagraph{BYOL without Batch Normalization Layer.} The widely known fact about BYOL~\cite{grill2020bootstrap} is that this method falls into the mode collapse~\cite{untitled-ai} without batch normalization layer. The authors of~\cite{grill2020bootstrap} performed studies that BYOL works even without BN layer~\cite{richemond2020byol_wo_bn}. In this paper, authors showed that BYOL without BN gets matched performance using various normalization techniques including weight standardization~\cite{weight_standardization} or the deliberately handled initialization. But still, BYOL fails to converge optimal solution with such deliberately tuned training techniques. In this section, we show that \ours also work with layer normalization~\cite{layer_norm} without any other techniques in Table~\ref{table:repulsive_coeff_effect_layernorm}.  

In Table~\ref{table:repulsive_coeff_effect_layernorm}, the top-1 classification accuracy is largely degraded from 89.5\% to 10.6\%, i.e. mode collapsed. Ours with the Brownian diffusive loss $\mathcal{L}_b$ was not the case (compare 2nd vs. 6th row). Though we observe a slight degradation in the top-1 classification accuracy, ours sufficiently avoid collapsed representations. Further, we evaluate the BYOL with our Brownian diffusive loss to demonstrate its effectiveness against a mode collapse. We observe that our Brownian diffusive loss helps avoid collapsed representations (compare 3rd vs. 4th rows). We also observe that the quality of representations depends on the strength of the hyperparameter $\lambda_b$ where we obtain the best performance with $\lambda_b=5\times 10^{-4}$. We observe a tension as we see a smaller or larger $\lambda_b$ slightly degrades the quality of representations.

\begin{table}
  \caption{Comparison of the quality of representations between BYOL~\cite{grill2020bootstrap} and ours on the STL-10 dataset~\cite{stl10}. The Top-1 classification accuracy is reported with different types of normalization techniques: a batch normalization (BN)~\cite{batch_norm} and a layer norm (LN)~\cite{layer_norm}. To see the effect of our proposed Brownian Diffusive Loss, $\mathcal{L}_b$, we also report scores of BYOL with $\mathcal{L}_b$ (4th row).}
  \label{table:repulsive_coeff_effect_layernorm}
  \centering
  \begin{tabular}{lcccc}
    \toprule 
    Method & Norm. Layer & Batch Size & $\lambda_b$ & Top-1 (\%)\\ \midrule
    BYOL & BN & 256 & 0 & 89.5 \\
    Ours & BN & 256 & $5\times 10^{-2}$ & 91.4 \\\midrule
    BYOL & LN & 256 & 0 & 10.6 \\
    BYOL + our $\mathcal{L}_b$ & LN & 256 & $5\times 10^{-3}$ & 75.3\\ 
    BYOL & LN & 1024 & 0 & 10.6 \\\midrule
    Ours & LN & 256 & $5\times 10^{-4}$ & 80.7 \\
    Ours & LN & 256 & $5\times 10^{-3}$ & 82.3 \\
    Ours & LN & 256 & $5\times 10^{-2}$ & 78.7 \\
    \bottomrule
  \end{tabular}
  \vspace{1em}
\end{table}

\clearpage
%
%
\bibliographystyle{splncs04}
\bibliography{egbib}
\end{document}